\documentclass[letterpaper, 10 pt, conference]{IEEEtran}
\IEEEoverridecommandlockouts                          
\usepackage{graphics} 
\usepackage{epsfig}
\usepackage{times}
\usepackage{amsmath}
\usepackage{amssymb}
\usepackage{threeparttable}
\usepackage{booktabs}
\usepackage{soul}
\usepackage{subfigure}
\usepackage{xcolor}
\usepackage[hidelinks]{hyperref}
\usepackage{algorithm}
\usepackage{algorithmic,comment}
\usepackage{tcolorbox}

\title{\LARGE \bf
Adaptive Motion Planning via Contact-Based Intent Inference for Human-Robot Collaboration}

\author{Jiurun Song$^{1}$, Xiao Liang$^{2}$, and Minghui Zheng$^{3}$
\thanks{*This work was supported by the USA National Science Foundation under Grant No. 2026533/2422826.}
\thanks{$^{1}$Department of Mechanical Engineering, Texas A\&M University, College Station, TX 77843, USA {\tt\small jiurun\_song@tamu.edu}}
\thanks{$^{2}$Department of Civil and Environmental Engineering, Texas A\&M University, College Station, TX 77843, USA {\tt\small xliang@tamu.edu}}
\thanks{$^{3}$Department of Mechanical Engineering, Texas A\&M University, College Station, TX 77843, USA {\tt\small mhzheng@tamu.edu}}}

\begin{document}

\maketitle
\thispagestyle{empty}
\pagestyle{empty}

\begin{abstract}

Human-robot collaboration (HRC) requires robots to adapt their motions to human intent to ensure safe and efficient cooperation in shared spaces. Although large language models (LLMs) provide high-level reasoning for inferring human intent, their application to reliable motion planning in HRC remains challenging. Physical human-robot interaction (pHRI) is intuitive but often relies on continuous kinesthetic guidance, which imposes burdens on operators. To address these challenges, a contact-informed adaptive motion-planning framework is introduced to infer human intent directly from physical contact and employ the inferred intent for online motion correction in HRC. First, an optimization-based force estimation method is proposed to infer human-intended contact forces and locations from joint torque measurements and a robot dynamics model, thereby reducing cost and installation complexity while enabling whole-body sensitivity. Then, a torque-based contact detection mechanism with link-level localization is introduced to reduce the optimization search space and to enable real-time estimation. Subsequently, a contact-informed adaptive motion planner is developed to infer human intent from contacts and to replan robot motion online, while maintaining smoothness and adapting to human corrections. Finally, experiments on a 7-DOF manipulator are conducted to demonstrate the accuracy of the proposed force estimation method and the effectiveness of the contact-informed adaptive motion planner under perception uncertainty in HRC.

\end{abstract}

\section{Introduction}

\begin{figure}[!t]
    \centering
    \includegraphics[width=0.95\linewidth]{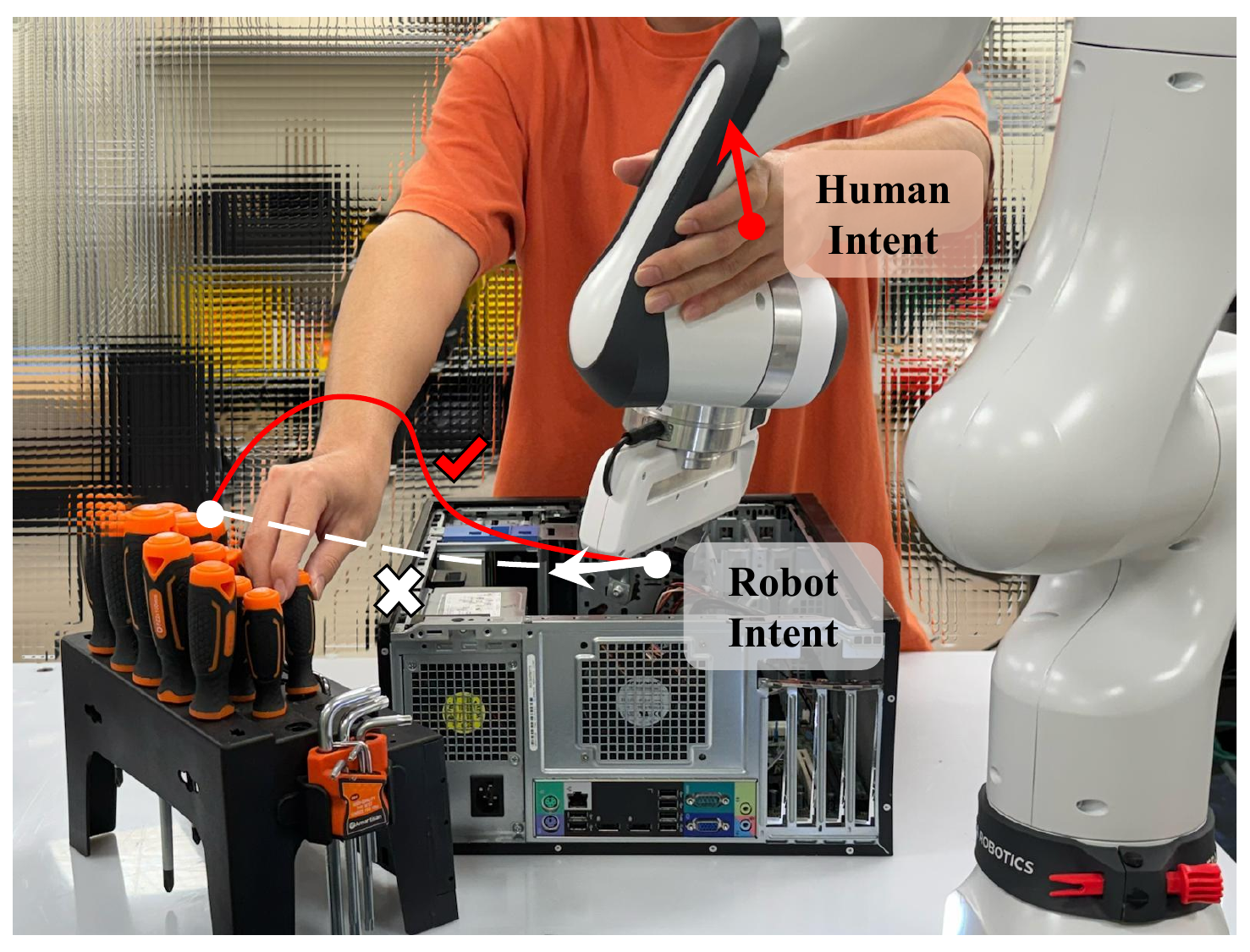}
    \caption{Human-robot collaboration in a shared space. Although the ultimate goals are aligned, the robot’s planned path does not account for the human arm as an obstacle, while the human perceives the collision risk and attempts to correct the robot’s motion through physical interaction to ensure safety.}
    \label{fig:Introduction}
\end{figure}

Human-robot collaboration (HRC) has gained increasing attention with the growing deployment of robots in unpredictable environment~\cite{ajoudani2018progress}. Advantages in safety, flexibility, and efficiency are realized by integrating robotic precision with human dexterity, thereby enabling tasks such as disassembly~\cite{liu2024kg,liu2025raiserobot,liu2024hybrid}, polishing ~\cite{gaz2018model,kana2021impedance}, and rehabilitation training~\cite{zhou2021human,zhang2024actuator}. In these applications, the fundamental objective is the assurance of safe and efficient interaction with humans in a shared workspace. Although the ultimate goals of humans and robots are generally aligned, differences in motion planning may arise during task execution in uncertain and dynamic environments, as shown in Fig.~\ref{fig:Introduction}. Therefore, accurate and timely adjustment of robot motion to human intent is considered essential for transforming robots from simple tools into intuitive and trustworthy collaborative partners in HRC.

\subsection{Related Work}

Large language models (LLMs) have been regarded as effective tools for interpreting human intent in HRC~\cite{ahn2022can,wu2023tidybot,lynch2023interactive}. By processing written or spoken commands, LLMs can decompose human input into a sequence of high-level actions, which makes them particularly effective for task-level planning~\cite{ao2025llm,zhou2024isr}. However, many HRC applications, such as disassembly and polishing, require continuous low-level motion adaptation rather than discrete action specification~\cite{liu2023task}. In these scenarios, natural language commands are insufficient for conveying the precise magnitude and direction of corrections, and may also introduce execution delays~\cite{lynch2023interactive}, incomplete descriptions~\cite{cui2023no,shi2024yell}, and errors in long-horizon tasks~\cite{zha2024distilling,jiang2024transic}. Therefore, although LLMs are powerful in semantic reasoning and task-level decision making, they remain inadequate for reliable motion planning in uncertain and dynamic environments.

Physical human-robot interaction (pHRI) provides an intuitive and adaptable interface for HRC, enabling humans to convey intent and directly influence robot motion at the low-level control stage~\cite{de2008atlas,haddadin2016physical,farajtabar2024path}. The central challenge lies in accurately interpreting pHRI so that robots can adapt their motions to generate quick, safe, and human-aligned responses. A common approach is kinesthetic guidance, in which a human physically pushes, pulls, or twists a compliant robot along a desired trajectory, effectively acting as an expert demonstrator~\cite{jarrasse2012framework,mortl2012role}. In this mode, the robot yields to external forces without interpreting their meaning, and typically resumes its original trajectory once the interaction ends, thereby treating human corrections as disturbances rather than as informative cues. However, physical interventions often reflect human intent and can serve as demonstrations of desired motion~\cite{losey2018review,hoffman2024inferring}. Building on this insight, prior research has used short episodes of kinesthetic guidance to infer corrective intent and update task objectives, allowing robots to adjust their motions consistent with human intent even after the interaction concludes~\cite{losey2022physical,li2021learning,mehta2024unified}. In~\cite{losey2022physical}, human intent is inferred by modeling physical interaction as observations of a latent objective, with real-time updates enabling adaptive motion. Nevertheless, real-world objective functions are often high-dimensional and cannot be fully conveyed through limited corrections. To address this, Xie et al.~\cite{xie2024safe} proposed a certifiable framework that refines robot motions by learning implicit safety constraints from online human directional feedback and embedding them into model predictive control. However, this approach remains limited by the task dependence of the learned constraints and its reliance on frequent human corrections.

Another challenge in pHRI arises from the interaction modality. Although kinesthetic corrections provide immediate and continuous feedback that reflects human-intended adjustments, they require sustained physical effort and thus impose a considerable burden on the operator. In contrast, contact force information can directly convey intent, as both magnitude and direction inherently encode the desired modification of motion~\cite{losey2017trajectory}. However, accurately measuring or estimating these forces remains challenging. Explicit tactile sensors and artificial skins offer high sensitivity and precision for perceiving physical contacts~\cite{cheng2019comprehensive,mittendorfer2011humanoid,armleder2022interactive}, but practical deployment is hindered by challenges such as conformity to curved surfaces, wiring complexity, power supply, and robustness under impacts. Alternatively, integrated joint torque sensors combined with model-based dynamics observers are widely used to estimate external forces~\cite{magrini2014estimation,manuelli2016localizing,zwiener2018contact}, yet most approaches are limited in real-time capability and are typically restricted to regions near the end-effector.

\subsection{Contributions}

In HRC, constrained and uncertain workspaces often result in incomplete environmental perception, where vision occlusions introduce uncertainty and necessitate human feedback to adapt robot motion for safety and efficiency under dynamic changes~\cite{tian2023optimization,liu2024integrating}. As discussed above, when human intent and robot motion are misaligned, kinesthetic guidance at the low-level control stage provides an intuitive means of correction. However, conventional kinesthetic guidance requires sustained physical effort, imposing a significant burden on the operator. Since the magnitude and direction of external contact forces inherently encode human intent, such information can be exploited to reduce the need for continuous intervention. In this article, an adaptive motion planning framework is proposed to infer human intent directly from contact force estimation, enabling corrections to be conveyed through brief contacts rather than sustained kinesthetic guidance.

The main contributions are as follows:

\begin{enumerate}
    \item An optimization-based force-estimation method is developed to infer human-intended contact forces and locations from joint-torque sensors and a robot dynamics model, thereby reducing cost and installation complexity while enabling whole-body sensitivity to corrective contacts.
    
    \item A torque-based contact detection mechanism with link-level localization is introduced to reduce the solution space of the force-estimation optimization problem, thereby enabling real-time estimation of human-intended contact forces.
    
    \item A contact-informed adaptive motion planner is developed to interpret human intent from contact information and to replan trajectories in real time, ensuring smoothness while adapting to human corrections.
\end{enumerate}

The remainder of this article is organized as follows. Section~\ref{sec:preliminaries} introduces the problem preliminaries. Section~\ref{sec:method} presents the proposed methodology, including torque-based contact detection, link-level localization, optimization-based contact force estimation, and contact-informed adaptive motion planning. Section~\ref{sec:experiments} reports the experimental results on a 7-DOF robot manipulator. Finally, Section~\ref{sec:conclusion} concludes the work and outlines directions for future research.

\begin{figure*}[!t]
    \centering
    \includegraphics[width=0.95\linewidth]{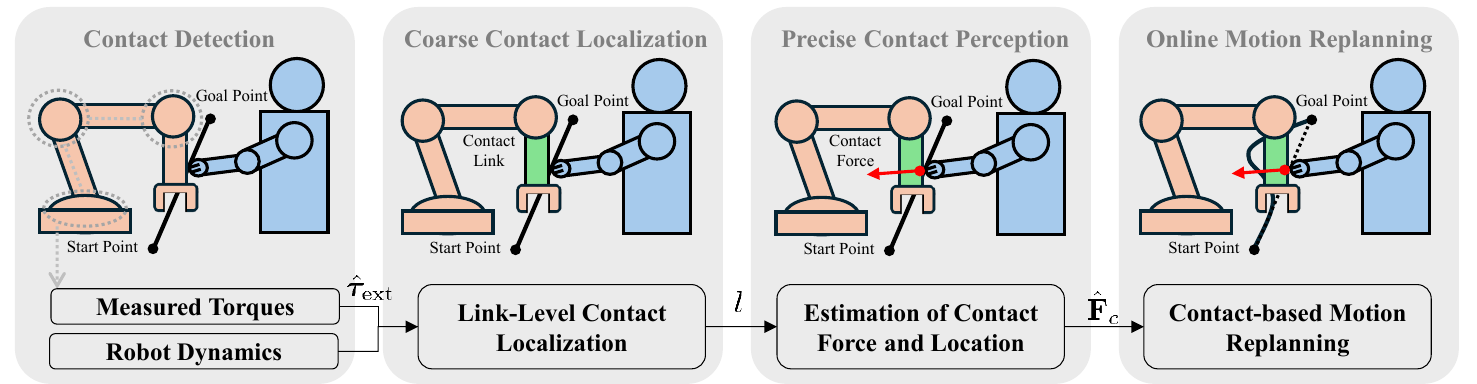}
    \caption{Contact is detected using torque measurements together with the robot dynamics model. Once a contact event is identified, the corresponding link is determined to reduce the search space for contact localization. Constrained optimization is then applied to the identified link to estimate the precise contact force and position. Finally, the estimated contact information is utilized to infer human intent and to adapt the robot motion accordingly.}
    \label{fig:Methodology}
\end{figure*}

\section{Problem Preliminaries}
\label{sec:preliminaries}

Consider an $n$-DOF serial manipulator with joint configuration $\mathbf{q} \in \mathbb{R}^n$, joint velocity $\dot{\mathbf{q}} \in \mathbb{R}^n$, joint acceleration $\ddot{\mathbf{q}} \in \mathbb{R}^n$, and actuation torque vector $\boldsymbol{\tau} \in \mathbb{R}^n$. The standard rigid-body dynamics neglecting external forces can be written as
\begin{equation}
\boldsymbol{\tau} \;=\; \mathbf{M}(\mathbf{q})\,\ddot{\mathbf{q}} \;+\; \mathbf{C}(\mathbf{q},\dot{\mathbf{q}})\,\dot{\mathbf{q}} \;+\; \mathbf{G}(\mathbf{q}) \;+\; \mathbf{F}_{f}(\dot{\mathbf{q}}),
\label{eq:dynamics}
\end{equation}
where $\mathbf{M}(\mathbf{q}) \in \mathbb{R}^{n \times n}$ is the symmetric positive-definite inertia matrix, $\mathbf{C}(\mathbf{q},\dot{\mathbf{q}})\dot{\mathbf{q}} \in \mathbb{R}^n$ is the vector of Coriolis and centripetal terms, $\mathbf{G}(\mathbf{q}) \in \mathbb{R}^n$ is the gravity term, and $\mathbf{F}_{f}(\dot{\mathbf{q}}) \in \mathbb{R}^n$ represents joint friction. For simplicity, the nonlinear internal terms are grouped into a single vector $\mathbf{h}(\mathbf{q},\dot{\mathbf{q}}) = \mathbf{C}(\mathbf{q},\dot{\mathbf{q}})\,\dot{\mathbf{q}} + \mathbf{G}(\mathbf{q}) + \mathbf{F}_{f}(\dot{\mathbf{q}})$. Using this notation, the dynamics including the torques induced by external contact can be expressed as:
\begin{equation}
\boldsymbol{\tau} \;=\; \mathbf{M}(\mathbf{q})\,\ddot{\mathbf{q}} \;+\; \mathbf{h}(\mathbf{q},\dot{\mathbf{q}}) \;+\; \boldsymbol{\tau}_{\mathrm{ext}},
\label{eq:extDynamics}
\end{equation}
where $\boldsymbol{\tau}_{\mathrm{ext}} \in \mathbb{R}^n$ denotes the generalized joint torques arising from external contact forces.

To derive the relationship between a contact force and the corresponding joint torques, the principle of virtual work is applied. Consider an external contact force $\mathbf{F}_c \in \mathbb{R}^3$ applied at a point on the manipulator, with an associated virtual displacement $\delta \mathbf{x}_c \in \mathbb{R}^3$. The virtual work done by this force is 
\begin{equation}
\delta W \;=\; \mathbf{F}_c^\mathsf{T}\,\delta \mathbf{x}_c,
\label{eq:virtWork}
\end{equation}
Kinematically, the virtual displacement of the contact point is related to a virtual joint displacement $\delta \boldsymbol{\theta} \in \mathbb{R}^n$ by the Jacobian matrix $J_c(\mathbf{q}) \in \mathbb{R}^{3 \times n}$:
\begin{equation}
\delta \mathbf{x}_c \;=\; J_c(\mathbf{q})\,\delta \boldsymbol{\theta}.
\label{eq:deltaX}
\end{equation}
Substituting \eqref{eq:deltaX} into \eqref{eq:virtWork} gives 
\begin{equation}
\delta W \;=\; \mathbf{F}_c^\mathsf{T}\,J_c(\mathbf{q})\,\delta \boldsymbol{\theta}.
\label{eq:virtWorkJ}
\end{equation}
By the principle of virtual work, this must equal the work done by the corresponding generalized joint torques $\boldsymbol{\tau}_{\mathrm{ext}}$ for the same virtual motion:
\begin{equation}
\delta W \;=\; \boldsymbol{\tau}_{\mathrm{ext}}^\mathsf{T}\,\delta \boldsymbol{\theta}.
\label{eq:virtWorkTau}
\end{equation}
Since $\delta \boldsymbol{\theta}$ is arbitrary, it follows that 
\begin{equation}
\boldsymbol{\tau}_{\mathrm{ext}} \;=\; J_c(\mathbf{q})^\mathsf{T}\,\mathbf{F}_c,
\label{eq:tauExtRel}
\end{equation}
which provides the fundamental relationship between a contact force and the equivalent joint torques via the Jacobian transpose.

Given the robot dynamic model and the mapping between contact forces and joint torques, the problem is formulated as the estimation of human-intended contact forces and their locations from residual torques ${\tau}_{\mathrm{ext}}$, and the exploitation of the force magnitude and direction to infer human intent for adaptive motion replanning in HRC.

\section{Contact-Informed Adaptive Motion Planning}
\label{sec:method}

An contact-informed adaptive motion planning framework is proposed to enable robot motion adaptation through physical human-robot interaction. The framework integrates torque-based contact detection, link-level localization, and optimization-based force estimation to infer human intent from contact forces. A contact-informed motion planner then replans motions while ensuring smoothness and continuity. The overall workflow is shown in Fig.~\ref{fig:Methodology}, and the details of each module are presented in the following subsections.

\subsection{Torque-based Contact Detection}

In~\eqref{eq:tauExtRel}, the joint torque vector $\boldsymbol{\tau}{\mathrm{ext}}$ induced by an external contact force is expressed in terms of the unknown force and the Jacobian $J_c(\mathbf{q})$, which depends on the robot state and the contact position. If $\boldsymbol{\tau}{\mathrm{ext}}$ were available, both the magnitude and location of the contact force could be estimated by solving an optimization problem. In practice, however, $\boldsymbol{\tau}_{\mathrm{ext}}$ is not directly measurable and can be inferred from the difference between measured joint torques and model-predicted torques obtained from the robot dynamics in~\eqref{eq:dynamics}. This motivates the development of a torque-based contact detection method that identifies physical interactions by monitoring residual torques.

Most commercial robotic manipulators are equipped with joint torque sensors that provide the measured torque $\tau_{\mathrm{meas},j}(t)$ at each joint $j$. For the same joint, the model-predicted torque $\tau_{\mathrm{model},j}(t)$ can be computed using~\eqref{eq:dynamics} and the current robot state. The residual torque at joint $j$ is defined as $\hat{\tau}_{\mathrm{ext},j}(t) = \tau_{\mathrm{meas},j}(t) - \tau_{\mathrm{model},j}(t)$. In vector form, the joint-space residual torque can be written as
\begin{equation}
\hat{\boldsymbol{\tau}}_{\mathrm{ext}}(t) \;=\; \boldsymbol{\tau}_{\mathrm{meas}}(t) \;-\; \boldsymbol{\tau}_{\mathrm{model}}(t)\,,
\label{eq:residual}
\end{equation}
where $\hat{\boldsymbol{\tau}}_{\mathrm{ext}}(t) \in \mathbb{R}^n$ represents the residual torque vector, which serves as an estimate of $\boldsymbol{\tau}_{\mathrm{ext}}$ based on sensor measurements.

To detect external contact, a scalar detection statistic is derived from the residual torque vector using a weighted $\ell_2$ norm, which provides a unified measure of the residual magnitude across all joints. Although joint torque sensors are initially calibrated to the same nominal sensitivity and accuracy, deviations may arise over time due to wear, aging, or operating conditions. The weighting matrix compensates for these deviations to improve contact detection robustness. The detection statistic is defined as
\begin{equation}
\eta(t) \;=\; \Big\|\, \mathbf{W}_{\tau}\,\hat{\boldsymbol{\tau}}_{\mathrm{ext}}(t) \,\Big\|_2 ,
\label{eq:eta}
\end{equation}
where $\mathbf{W}_{\tau} \in \mathbb{R}^{n\times n}$ is a diagonal weighting matrix and $\|\cdot\|_2$ denotes the Euclidean norm. The scalar $\eta(t)$ quantifies the overall magnitude of the residual torque at time $t$.

Due to measurement noise and modeling errors, the detection statistic $\eta(t)$ may fluctuate even during free motion. To reduce these fluctuations, an exponentially weighted moving average (EWMA) is applied as a first-order low-pass filter:  
\begin{equation}
\bar{\eta}(k) \;=\; \alpha\,\eta(k) \;+\; (1-\alpha)\,\bar{\eta}(k-1)\,,
\label{eq:ewma}
\end{equation}
where $0<\alpha\leq 1$ is the forgetting factor. 

A contact is detected when $\bar{\eta}(k)$ exceeds a predefined threshold consistently over several consecutive samples. To ensure reliable state transitions and to avoid chattering, a hysteresis mechanism is incorporated. Let $\theta_{\tau} > 0$ denote the detection threshold, and let $N_{\mathrm{on}}, N_{\mathrm{off}} \in \mathbb{N}_{+}$ denote the required counts of consecutive samples to switch the state on and off, respectively. The contact state $C(k)$ is defined such that $C(k)=1$ indicates the presence of contact and $C(k)=0$ indicates no contact. The state is updated according to  
\begin{equation}
\label{eq:hysteresis}
C(k) \;=\;
\begin{cases}
1, & \displaystyle \min_{\,i\in\{k-N_{\mathrm{on}}+1,\ldots,k\}} \bar{\eta}(i) \;\ge\; \theta_{\tau},\\[6pt]
0, & \displaystyle \max_{\,i\in\{k-N_{\mathrm{off}}+1,\ldots,k\}} \bar{\eta}(i) \;\le\; \theta_{\tau},\\[2pt]
C(k-1), & \text{otherwise.}
\end{cases}
\end{equation}
with the initial condition $C(0)=0$. The state switches to one when $\bar{\eta}(k)$ has exceeded the threshold for $N_{\mathrm{on}}$ consecutive samples, and returns to zero when it has remained below the threshold for $N_{\mathrm{off}}$ consecutive samples. If neither condition is satisfied, the state retains its previous value.

\subsection{Link-Level Contact Localization}

To narrow the search space for contact force estimation, a coarse localization strategy is employed. After detecting a contact, the corresponding link is identified as the approximate contact position, which reduces the dimensionality of the optimization and improves computational efficiency and robustness.

An external contact force applied to the robot induces residual torques in the joints between the base and the contact location, whereas joints beyond the contact point, closer to the end effector, are subject to little or no disturbance since the force is not propagated further. To determine whether a residual torque is relevant for localization, a threshold 
$\tau_{\mathrm{th}}>0$ is introduced. A residual torque at joint $j$ is considered \textit{meaningful} when $|\hat{\tau}_{\mathrm{ext},\,j}| > \tau_{\mathrm{th}}$, and \textit{negligible} otherwise. This criterion is employed to localize the contact by analyzing the distribution of $\hat{\tau}_{\mathrm{ext},j}$ along the manipulator. Starting from the base and moving outward, the index of the contacted link $\ell$ is 
identified as the last joint for which the residual torque is significant, whereas the subsequent joint is not. Formally, $\ell$ is determined as
\begin{equation}
\label{eq:linkIdentify}
\ell = \max \Big\{\, j \;\Big|\; |\hat{\tau}_{\mathrm{ext},\,j}| > \tau_{\mathrm{th}} \;\text{and}\; |\hat{\tau}_{\mathrm{ext},\,j+1}| < \tau_{\mathrm{th}} \,\Big\}\,,
\end{equation}
with the convention that $\ell=n$ if $|\hat{\tau}_{\mathrm{ext},n}| > \tau_{\mathrm{th}}$, meaning that joints $1$ through $\ell$ exhibit meaningful residual torque, whereas joint $\ell+1$ does not, thereby identifying link $\ell$ as the contacted link.

\subsection{Contact Estimation via Constrained Optimization}

Building on contact detection and link identification, the contact force and its precise location are estimated through a constrained optimization problem, enabling appropriate robot response to physical interaction.

The contact is modeled as a point contact on link $\ell$, which is appropriate when the contact area is small relative to the link geometry and the net effect can be represented by a single resultant force. Each link is abstracted by a centerline curve $\boldsymbol{\gamma}_\ell:[0,1]\!\to\!\mathbb{R}^3$, while the link’s diameter and cross-sectional geometry are neglected. The contact location is parameterized by a normalized arc length $u\in[0,1]$, where $u=0$ corresponds to the proximal joint $\ell$ and $u=1$ to the distal joint $\ell+1$. When the curvature of $\boldsymbol{\gamma}_\ell$ is negligible, the centerline is approximated by the straight segment connecting the base and tip of the link. Denoting these endpoints by $\mathbf{p}_{\ell,\mathrm{base}}\in\mathbb{R}^3$ and $\mathbf{p}_{\ell,\mathrm{tip}}\in\mathbb{R}^3$ in the robot’s base frame, the Cartesian position of a candidate contact point is expressed as 
\begin{equation}
\mathbf{p}_c(u) \;=\; \mathbf{p}_{\ell,\mathrm{base}} \;+\; u\big(\mathbf{p}_{\ell,\mathrm{tip}} - \mathbf{p}_{\ell,\mathrm{base}}\big), 
\label{eq:pc}
\end{equation}
with $0 \le u \le 1$. It provides a continuous parameterization of the contact location along the link centerline, thereby reducing the dimensionality of the estimation problem and improving computational efficiency.

The unknown parameters to be estimated are the contact location $s$ and the contact force vector $\mathbf{F}_c \in \mathbb{R}^3$. An estimate of these can be obtained by solving a nonlinear least-squares problem that fits the predicted external joint torques to the measured residuals. Suppose a contact persists over a time interval during which $N$ samples of 
the external residual torque are collected. Denote these measurements as $\hat{\boldsymbol{\tau}}_{\mathrm{ext}}(t_k)$ for $k = 1, 2, \dots, N$, using the notation from~\eqref{eq:residual}. The Cartesian position of the candidate contact point on link $\ell$ is parameterized by $\mathbf{p}_c(s)$ as given in~\eqref{eq:pc}. The Jacobian that maps a force at $\mathbf{p}_c(s)$ to the corresponding joint torques is defined as
\begin{equation}
J_c(\mathbf{q}(t_k),\,s) \;=\; \frac{\partial \mathbf{p}_c(s)}{\partial \mathbf{q}(t_k)} \,\in\, \mathbb{R}^{3 \times n}\,,
\label{eq:Jc}
\end{equation}
where $\mathbf{q}(t_k)$ is the joint configuration at time $t_k$. For any candidate contact location $s$ and force vector $\mathbf{F}_c$, the predicted external joint torque at time $t_k$ is given by $J_c(\mathbf{q}(t_k),\,s)^\mathsf{T}\mathbf{F}_c$. The discrepancy between this prediction and the measured residual $\hat{\boldsymbol{\tau}}_{\mathrm{ext}}(t_k)$ indicates how well the candidate pair $(s,\mathbf{F}_c)$ explains the observed contact effect. Therefore, a cost function is defined as the sum of squared errors over all samples:
\begin{equation}
f(s,\,\mathbf{F}_c) \;=\; \tfrac{1}{2}\,\sum_{k=1}^{N} 
\Big\|\,\hat{\boldsymbol{\tau}}_{\mathrm{ext}}(t_k)\;-\;
J_c\!\big(\mathbf{q}(t_k),\,s\big)^\mathsf{T}\,\mathbf{F}_c \,\Big\|^2\,,
\label{eq:cost}
\end{equation}

Accordingly, the contact force estimation problem is formulated as the constrained nonlinear optimization:
\begin{equation}
\begin{aligned}
\min_{\,s,\,\mathbf{F}_c}\quad & \tfrac{1}{2}\sum_{k=1}^{N} \Big\|\,\hat{\boldsymbol{\tau}}_{\mathrm{ext}}(t_k)\;-\;J_c\!\big(\mathbf{q}(t_k),\,s\big)^\mathsf{T}\,\mathbf{F}_c \,\Big\|^2, \\
\text{s.t.}\quad & 0 \;\le\; s \;\le\; 1, \\
& \|\mathbf{F}_c\| \;\le\; F_c^{\max}\,,
\end{aligned}
\label{eq:opt}
\end{equation}
where $F_c^{\max} > 0$ is the upper bound of the contact force magnitude. The first constraint confines the contact location to link $\ell$.

Although the optimization in~\eqref{eq:opt} is non-convex due to the nonlinear dependence of $J_c$ on $s$, the problem structure allows an efficient solution. For a fixed $s$, the cost function in~\eqref{eq:opt} is quadratic in $\mathbf{F}_c$ and has the closed-form least-squares solution
\begin{equation}
\mathbf{F}_c^{\star}(s) \;=\; \big(J_c^\mathsf{T} J_c + \lambda^2 I_3\big)^{-1} J_c^\mathsf{T} \,\hat{\boldsymbol{\tau}}_{\mathrm{ext}}\,,
\label{eq:Fstar}
\end{equation}
where $\lambda>0$ is a damping factor that improves numerical stability near singular configurations. The solution $\mathbf{F}_c^{\star}(s)$ is further projected onto the admissible set $\|\mathbf{F}_c\|\leq F_c^{\max}$ to satisfy the magnitude constraint. Substituting~\eqref{eq:Fstar} into~\eqref{eq:opt} reduces the problem to a one-dimensional search over $s$:
\begin{equation}
f(s) \;=\; \tfrac{1}{2}\,\big\|\hat{\boldsymbol{\tau}}_{\mathrm{ext}} - J_c(\mathbf{q},s)^\mathsf{T}\,\mathbf{F}_c^{\star}(s)\big\|^2\,.
\label{eq:reduced}
\end{equation}

To solve~\eqref{eq:reduced}, $s$ is discretized along the link, and the residual cost $f(s)$ is evaluated on the grid. The best candidate is refined using Brent’s algorithm, yielding an efficient approximation of the global optimum. The result is denoted by $(\hat{s}, \hat{\mathbf{F}}_c)$, where $\hat{s}$ is the estimated contact location on link $\ell$ and $\hat{\mathbf{F}}_c$ the corresponding Cartesian contact force.

\begin{figure*}[!t]
    \centering
    \includegraphics[width=0.99\linewidth]{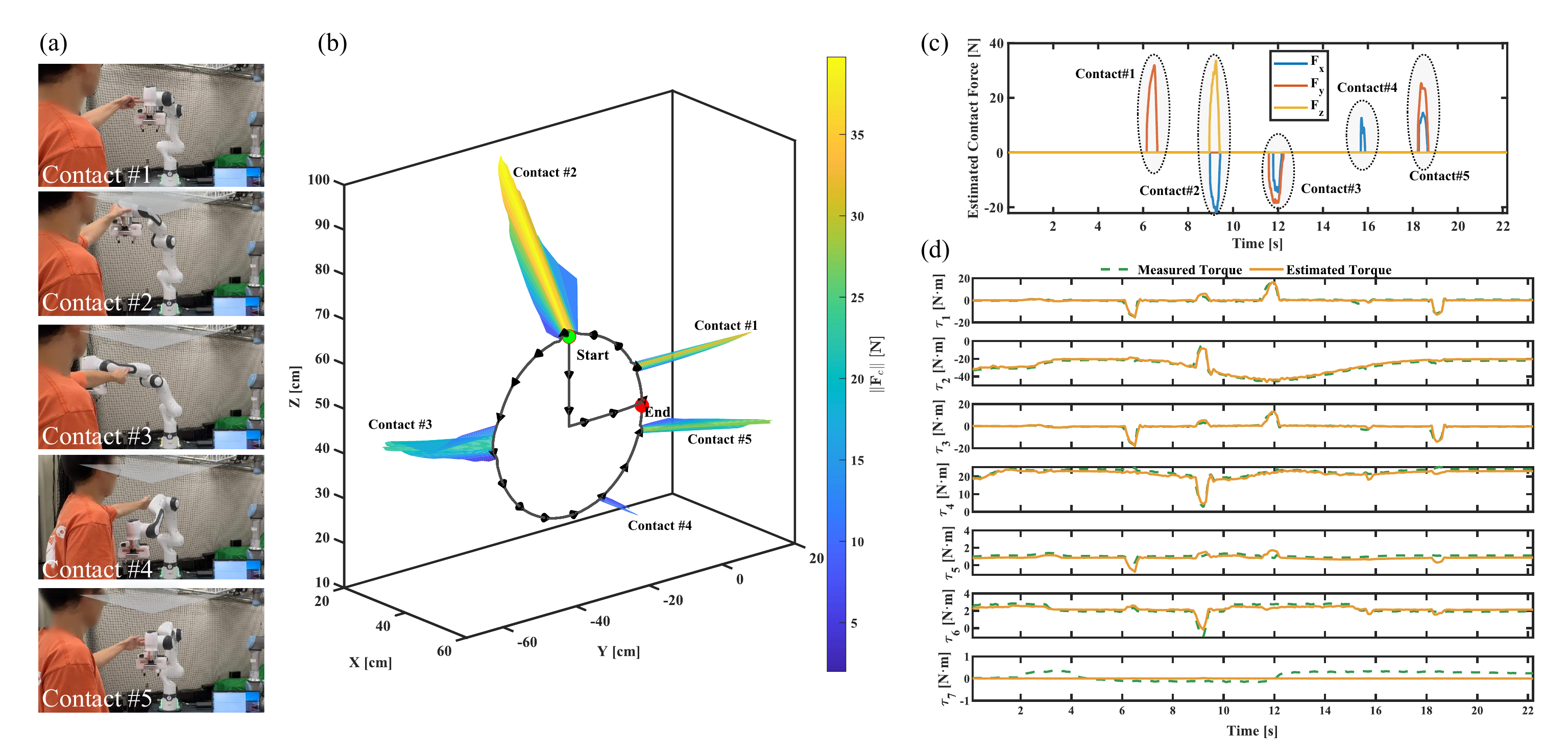}
    \caption{Force estimation results. (a) Experimental setup with human-applied pushes, pulls, and taps at different links during circular end-effector motion. (b) Estimated force directions and magnitudes along the circular path, with surfaces illustrating variations within each contact episode. (c) Time histories of three-dimensional force estimates. (d) Comparison of estimated and measured joint torques across seven joints.}
    \label{fig:Force_estimation}
\end{figure*}

\subsection{Contact-Informed Adaptive Motion Planner}

With the contact force and location estimated, the proposed planner adapts the robot motion online to account for human interaction. The strategy modifies the reference path incrementally based on the estimated force, enabling smooth adaptation and convergence to the intended goal.

Let $\mathbf{x}_d(s)$, $s\in[0,1]$, denote the original Cartesian path of the robot end-effector, with $s=0$ at the start and $s=1$ at the goal. The executed path is modeled as a deformed version of the original path, defined as
\begin{equation}
\tilde{\mathbf{x}}(s) \;=\; \mathbf{x}_d(s) \;+\; \mathbf{d}(s)\,,
\label{eq:deformedTraj}
\end{equation}
where $\mathbf{d}(s)$ is the deformation applied to the original path at position $s$. The deformation is zero at both the start and the goal, ensuring that the initial and final configurations remain unchanged.

To implement online motion adaptation, time is divided into sequential detection windows of fixed length. The $r$-th window, containing $N_d$ samples and ending at time index $k_r$, may include variations in the human contact force. For planner updates, a representative measure is extracted as the average contact force vector in window $r$, denoted $\bar{\mathbf{F}}_c^{(r)}$, along with its unit direction vector $\hat{\mathbf{f}}^{(r)}$:
\begin{equation}
\bar{\mathbf{F}}_c^{(r)} \;=\; \frac{1}{N_d}\sum_{j=k_r-N_d+1}^{\,k_r}\mathbf{F}_c(j)\,, 
\qquad 
\hat{\mathbf{f}}^{(r)} \;=\; \frac{\bar{\mathbf{F}}_c^{(r)}}{\big\|\bar{\mathbf{F}}_c^{(r)}\big\|}\,,
\label{eq:avgForce}
\end{equation}
where $\mathbf{F}_c(j)$ is the estimated contact force at sample $j$. The average force in window $r$ is denoted $\bar{\mathbf{F}}_c^{(r)}$, with $\hat{\mathbf{f}}^{(r)}$ as its unit direction vector. The averaging assumes that the intended human action remains approximately constant within the short window, while instantaneous forces fluctuate due to sensor noise and natural variability.

Using the averaged force, a Cartesian deviation $\Delta \mathbf{x}_c^{(r)}$ is computed to guide the path correction, defined as
\begin{equation}
\Delta \mathbf{x}_c^{(r)} \;=\; \alpha \;\operatorname{sat}\!\Big(\big\|\bar{\mathbf{F}}_c^{(r)}\big\|,\,\mathrm{F}_c^{\mathrm{max}}\Big)\; \hat{\mathbf{f}}^{(r)}\,,
\label{eq:deltaTarget}
\end{equation}
where $\alpha > 0$ is a gain with units $[\alpha] = [\mathrm{m}]/[\mathrm{N}]$ mapping force magnitude to Cartesian displacement, and $\mathrm{F}_c^{\mathrm{max}} > 0$ is the admissible force threshold. The operator $\operatorname{sat}(x, \mathrm{F}_c^{\mathrm{max}})$ denotes the saturation function, which limits $|\bar{\mathbf{F}}_c^{(r)}|$ before scaling by $\alpha$ to ensure safety. The deviation $\Delta\mathbf{x}_c^{(r)}$ is aligned with the unit force direction $\hat{\mathbf{f}}^{(r)}$.

To avoid drift when the human intent remains unchanged across consecutive windows, only the incremental change relative to the previous target is applied. Let $\Delta \mathbf{x}_c^{(r-1)}$ be the deviation from the previous window, initialized as $\Delta \mathbf{x}_c^{(0)}=\mathbf{0}$ for $r=1$. The incremental change is defined as
\begin{equation}
\Delta^{\mathrm{inc}} \mathbf{x}_c^{(r)} \;=\; \Delta \mathbf{x}_c^{(r)} \;-\; \Delta \mathbf{x}_c^{(r-1)}\,.
\label{eq:incrementDelta}
\end{equation}
Small incremental changes are suppressed when their magnitude is below a deadband threshold $\varepsilon \ge 0$, defined as
\begin{equation}
\Delta^{\mathrm{inc}} \mathbf{x}_{c,\varepsilon}^{(r)} \;=\;
\begin{cases}
\mathbf{0}, & \big\|\Delta^{\mathrm{inc}} \mathbf{x}_c^{(r)}\big\| \,\le\, \varepsilon,\\[4pt]
\Delta^{\mathrm{inc}} \mathbf{x}_c^{(r)}, & \text{otherwise,}
\end{cases}
\label{eq:deadband}
\end{equation}
thus disregarding spurious updates from noise or minor force variations and ensuring stable motion adaptation.

The next step is to determine the portion of the future motion to be adapted. Let $s_{r+1} \in [0,1]$ be the path parameter at the end of window $r$, corresponding to the position at time $k_r$. The horizon length $H_r$, proportional to the averaged contact force magnitude, specifies the path segment influenced by the $r$-th deformation and is defined as
\begin{equation}
H_r \;=\; \beta \, \big\|\bar{\mathbf{F}}_c^{(r)}\big\|\,,
\label{eq:horizon}
\end{equation}
where $\beta > 0$ is a design parameter. A larger contact force yields a longer adaptation horizon. To ensure convergence, the horizon is truncated near the end of the path so that the deviation vanishes before reaching the target. The effective horizon length is defined as
\begin{equation}
H_r^\star \;=\; \min\Big\{\,H_r,\; 1 - s_{r+1}\,\Big\}\,,
\label{eq:truncatedHorizon}
\end{equation}
which ensures $s_{r+1}+H_r^\star \le 1$, i.e., the deformation vanishes before the path endpoint.

A normalized local coordinate $\xi_r(s)$ is introduced on $[s_{r+1},,s_{r+1}+H_r^\star]$ to parameterize the deformation progress along the path segment. It is defined by a saturating linear function as
\begin{equation}
\xi_r(s) \;=\; \operatorname{sat}_{[0,\,1]}\!\left(\frac{s - s_{r+1}}{H_r^\star}\right),
\label{eq:xiSat}
\end{equation}
which ensures $\xi_r(s_{r+1})=0$, $\xi_r(s_{r+1}+H_r^\star)=1$, and $\xi_r(s)\in[0,1]$. A $C^1$ polynomial bump function is then used to shape the deformation profile, vanishing at both ends with zero derivative at the boundaries to ensure smooth entry and exit. The function is defined as
\begin{equation}
b(\xi_r) \;=\; 16\,\xi_r^{2}\,(1-\xi_r)^{2}\,,
\label{eq:bump}
\end{equation}
which satisfies $b(\xi_r=0)=b(\xi_r=1)=0$, $b'(\xi_r=0)=b'(\xi_r=1)=0$, and attains $\max_{\xi_r \in [0,1]} b(\xi_r)=1$.

Using the bump function $b(\xi)$, the incremental deformation $\boldsymbol{\phi}_r(s)$ for window $r$ is defined such that its magnitude corresponds to the desired increment $\Delta^{\mathrm{inc}} \mathbf{x}_{c,\varepsilon}^{(r)}$ and it vanishes outside the interval $[\,s_{r+1},\,s_{r+1}+H_r^\star\,]$. The cumulative deformation after incorporating window $r$ is then expressed as
\begin{equation}
\mathbf{d}^{(r)}(s) \;=\;
\begin{cases}
\mathbf{d}^{(r-1)}(s), & s < s_{r+1},\\[4pt]
\mathbf{d}^{(r-1)}(s) \;+\; \boldsymbol{\phi}_r(s), & s \ge s_{r+1}\,,
\end{cases}
\label{eq:deformIncrement}
\end{equation}
with the initialization $\mathbf{d}^{(0)}(s) \equiv \mathbf{0}$. The function $\boldsymbol{\phi}_r(s)$ is nonzero only for $s_{r+1} \le s \le s_{r+1}+H_r^\star$ and is constructed as $\boldsymbol{\phi}_r(s) = \Delta^{\mathrm{inc}} \mathbf{x}_{c,\varepsilon}^{(r)} \, b(\xi_r(s))$, ensuring that the deformation smoothly grows from zero at $s_{r+1}$ to the full increment at the midpoint ($\xi_r=0.5$) and then decays back to zero at $s_{r+1}+H_r^\star$. Since $s_{r+1}+H_r^\star \le 1$ by definition, the deformation vanishes before the end of the path, thereby preserving $\mathbf{d}(1)=\mathbf{0}$ and ensuring that the adapted motion rejoins the original goal without residual offset. If the human intent remains unchanged across consecutive windows, then $\Delta^{\mathrm{inc}} \mathbf{x}_{c,\varepsilon}^{(r)}=\mathbf{0}$ and no new deformation is added; otherwise, only the discrepancy is smoothly accumulated over the finite horizon $H_r^\star$.

\begin{figure*}[!t]
    \centering
    \includegraphics[width=0.99\linewidth]{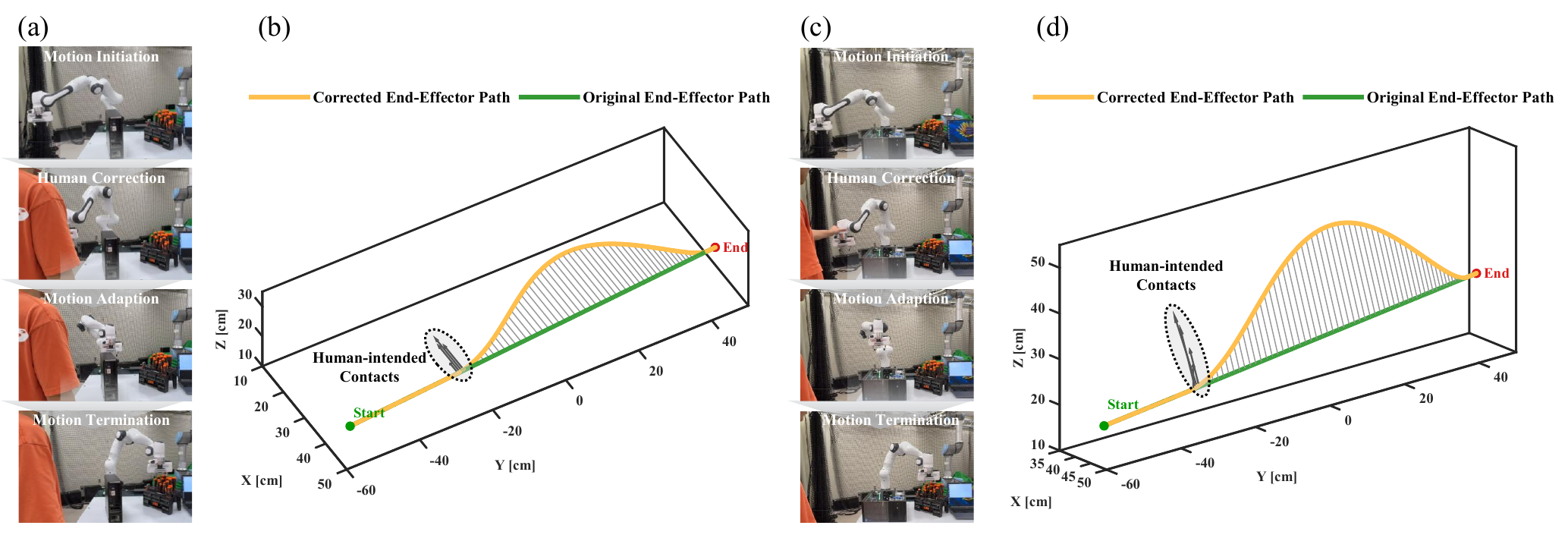}
    \caption{Experimental results of single-contact correction. (a) Snapshots and (b) corrected path with the original path, contact point, and estimated contact force in the first experiment; (c) and (d) present the corresponding results in the second experiment.}
    \label{fig:Single_contact_path}
\end{figure*}

\begin{figure}[!t]
    \centering
    \includegraphics[width=0.99\linewidth]{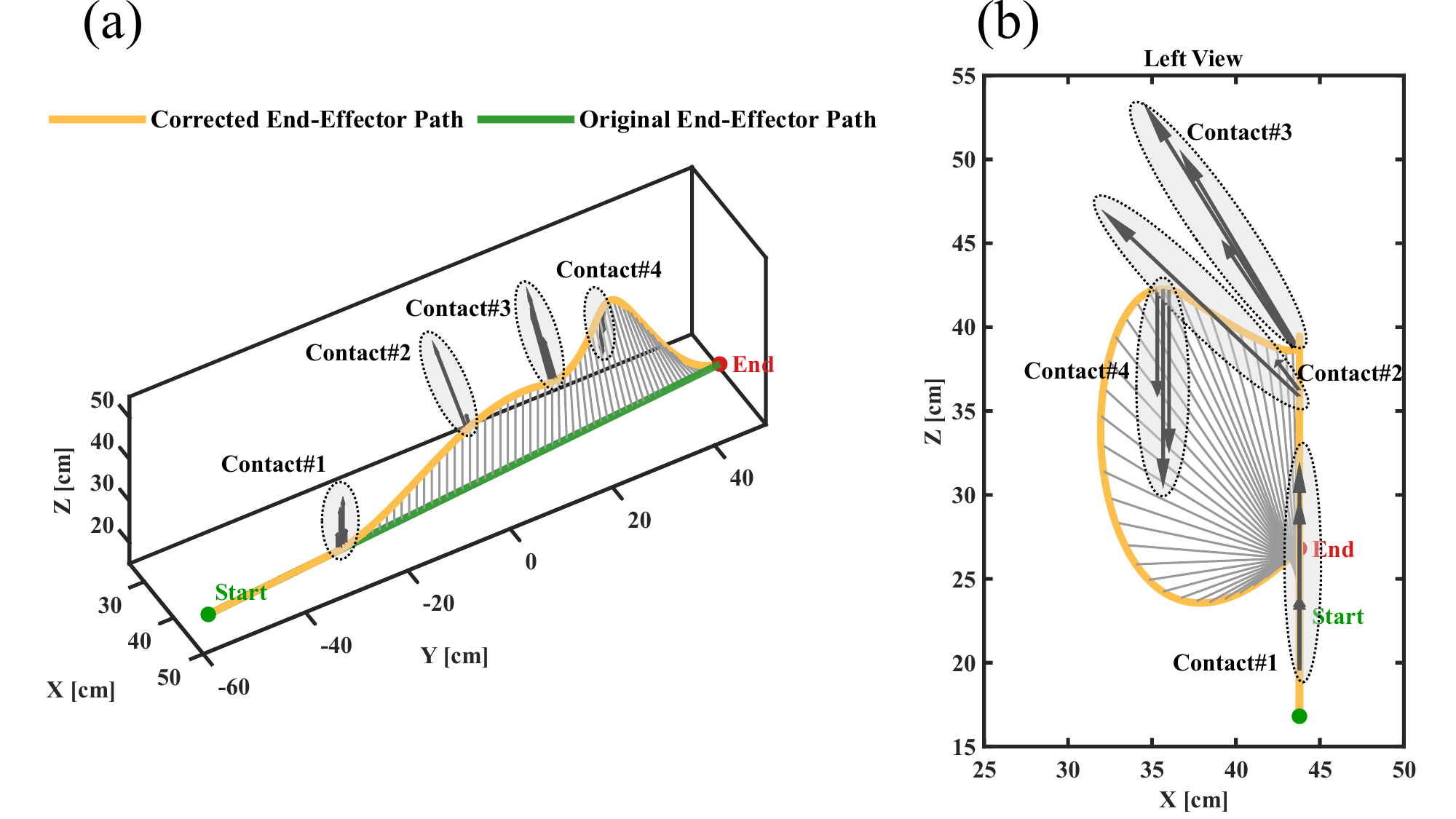}
    \caption{Experimental results of multi-contact correction. (a) Main view and (b) left view of the corrected path with the original path, contact points, and estimated contact forces in the third experiment.}
    \label{fig:Multi_contact_path}
\end{figure}

\section{Experiments}
\label{sec:experiments}

Experiments are conducted on a 7-DOF Franka Research 3 manipulator. The effectiveness of the proposed force estimation method is first validated, followed by adaptive motion planning results in a Human-robot collaboration scenario.

\subsection{Human-Intended Force Estimation}

The accuracy of the proposed force estimator was evaluated in a contact-only setting without motion adaptation. The robot executed a circular end-effector path as shown in Fig.~\ref{fig:Force_estimation}(b), while the estimator operated online. During execution, a human applied brief pushes, pulls, and taps at multiple links with varying directions, magnitudes, and durations, as illustrated in Fig.~\ref{fig:Force_estimation}(a).

Figure~\ref{fig:Force_estimation}(b) summarizes the estimated force directions and magnitudes along the circular path. For each contact episode, the estimated vectors span a surface to visualize force variation. The results show smaller forces at the beginning and end of each episode, peaking in the middle. Within an episode, force directions remain consistent, suggesting a single dominant intention.

Time histories of the three-dimensional force estimates are shown in Fig.~\ref{fig:Force_estimation}(c). Since direct installation of distributed force sensors is impractical, accuracy was evaluated indirectly by comparing joint torques predicted from the estimated forces with embedded torque measurements. As shown in Fig.~\ref{fig:Force_estimation}(d), the estimated torques follow the measured trends across all seven joints. Five prominent oscillations correspond to the five contact events, indicating timely detection. The mean absolute error of torque estimation is \(0.665\,\mathrm{N\cdot m}\).

\subsection{Human-Guided Adaptive Motion Planning}

The proposed adaptive motion planner was evaluated in a Human-robot collaboration scenario of computer disassembly. When the robot failed to detect an obstacle on its original path due to camera occlusion, the human conveyed intent through brief physical contacts, allowing the robot to avoid the obstacle and reach the goal safely.

\subsubsection{Single-contact correction} 

The first two experiments validated the effectiveness of the proposed method in avoiding potential collisions through single contacts. As shown in Fig.~\ref{fig:Single_contact_path}(a), the first experiment used an end-effector path defined as a straight line between the start and the goal, which was not detected to collide with a vertically placed computer case. Acting as a supervisor, the human observed the potential collision and applied a brief contact to adjust the motion. The estimated contact force was approximately horizontal along the negative $x$-axis, as shown in Fig.~\ref{fig:Single_contact_path}(b). The proposed planner interpreted this intent as an approximate lateral deviation and adapted the motion online. The corrected end-effector path smoothly bypassed the obstacle and reached the goal, as shown in Fig.~\ref{fig:Single_contact_path}(b). The second experiment considered a larger computer case placed horizontally, leaving no feasible lateral clearance. The human applied an approximately upward contact, as shown in Fig.~\ref{fig:Single_contact_path}(c), and the planner guided the robot to detour above the obstacle, resulting in a safe and effective path, as shown in Fig.~\ref{fig:Single_contact_path}(d).

\subsubsection{Multi-Contact Correction} 

Validation of the proposed method in avoiding potential collisions through multiple contacts is necessary, since the Human-robot collaboration environment is highly uncertain and a single contact may not always suffice. The third experiment repeated the first scenario but with an earlier initial correction, as humans often intervene in advance to ensure safety. After the first contact, the updated path still risked colliding with the vertical case. The operator therefore applied a second and a third contact to secure safe clearance, as shown in Fig.~\ref{fig:Multi_contact_path}(a). The third contact was intentionally strong, driving the path beyond the necessary deviation and delaying reconvergence. To improve efficiency, a subsequent downward contact accelerated the return toward the goal. The sequence of four contacts, their estimated directions and magnitudes, and the resulting corrected path compared with the original are presented in Fig.~\ref{fig:Multi_contact_path}(a) and Fig.~\ref{fig:Multi_contact_path}(b).

In summary, the experimental results demonstrate that the proposed method provides accurate real-time estimation of human-intended forces and enables adaptive motion planning that rapidly interprets human intent to adjust robot motion in HRC.

\section{Conclusion and Future Work}
\label{sec:conclusion}

A contact-informed adaptive motion-planning framework is proposed to infer human intent from physical contact and apply it for online motion correction in HRC. The proposed framework integrates torque-based contact detection, link-level localization, and optimization-based force estimation to identify where and how the human interacts with the robot, enabling online force estimation and achieving whole-body sensitivity. A contact-informed adaptive motion planner then interprets the inferred intent and adjusts robot motion in real time according to the estimated forces, providing smooth corrections without requiring continuous human guidance. Experiments on a 7-DOF manipulator validate the framework, showing that the estimated forces align with measured torques and that the robot successfully replans its path to avoid obstacles following brief human interactions in a collaborative disassembly task. The experimental results demonstrate that the proposed method enhances reliability and reduces human effort in shared tasks, leading to safer and more efficient HRC.  

Future work will focus on personalization and automation of the planner’s responsiveness. At present, sensitivity to contact forces is tuned manually and may not align with individual interaction styles. Learning-based approaches will be employed to adapt parameters from human feedback and iterative corrections across multiple trials, enabling user-specific force sensitivity. Such adaptations are expected to foster more intuitive and trustworthy Human-robot collaboration.

\addtolength{\textheight}{-1cm} 
\bibliographystyle{IEEEtran} 
\bibliography{refs}        

\begin{thebibliography}{10}
\providecommand{\url}[1]{#1}
\csname url@samestyle\endcsname
\providecommand{\newblock}{\relax}
\providecommand{\bibinfo}[2]{#2}
\providecommand{\BIBentrySTDinterwordspacing}{\spaceskip=0pt\relax}
\providecommand{\BIBentryALTinterwordstretchfactor}{4}
\providecommand{\BIBentryALTinterwordspacing}{\spaceskip=\fontdimen2\font plus
\BIBentryALTinterwordstretchfactor\fontdimen3\font minus \fontdimen4\font\relax}
\providecommand{\BIBforeignlanguage}[2]{{%
\expandafter\ifx\csname l@#1\endcsname\relax
\typeout{** WARNING: IEEEtran.bst: No hyphenation pattern has been}%
\typeout{** loaded for the language `#1'. Using the pattern for}%
\typeout{** the default language instead.}%
\else
\language=\csname l@#1\endcsname
\fi
#2}}
\providecommand{\BIBdecl}{\relax}
\BIBdecl

\bibitem{ajoudani2018progress}
A.~Ajoudani, A.~M. Zanchettin, S.~Ivaldi, A.~Albu-Sch{\"a}ffer, K.~Kosuge, and O.~Khatib, ``Progress and prospects of the human--robot collaboration,'' \emph{Autonomous Robots}, vol.~42, no.~5, pp. 957--975, 2018.

\bibitem{liu2024kg}
W.~Liu, K.~Eltouny, S.~Tian, X.~Liang, and M.~Zheng, ``Kg-planner: Knowledge-informed graph neural planning for collaborative manipulators,'' \emph{IEEE Transactions on Automation Science and Engineering}, vol.~22, pp. 5061--5071, 2024.

\bibitem{liu2025raiserobot}
C.~Liu, B.~Balasubramaniam, N.~Yancey, M.~Severson, A.~Shine, P.~Bove, B.~Li, X.~Liang, and M.~Zheng, ``Raise: A robot-assisted selective disassembly and sorting system for end-of-life phones,'' \emph{arXiv preprint arXiv:2509.23048}, 2025.

\bibitem{liu2024hybrid}
W.~Liu, C.~Liu, X.~Liang, and M.~Zheng, ``A hybrid task-constrained motion planning for collaborative robots in intelligent remanufacturing,'' \emph{Mechatronics}, vol. 102, p. 103222, 2024.

\bibitem{gaz2018model}
C.~Gaz, E.~Magrini, and A.~De~Luca, ``A model-based residual approach for human-robot collaboration during manual polishing operations,'' \emph{Mechatronics}, vol.~55, pp. 234--247, 2018.

\bibitem{kana2021impedance}
S.~Kana, S.~Lakshminarayanan, D.~M. Mohan, and D.~Campolo, ``Impedance controlled human--robot collaborative tooling for edge chamfering and polishing applications,'' \emph{Robotics and Computer-Integrated Manufacturing}, vol.~72, p. 102199, 2021.

\bibitem{zhou2021human}
J.~Zhou, Z.~Li, X.~Li, X.~Wang, and R.~Song, ``Human--robot cooperation control based on trajectory deformation algorithm for a lower limb rehabilitation robot,'' \emph{IEEE/ASME Transactions on Mechatronics}, vol.~26, no.~6, pp. 3128--3138, 2021.

\bibitem{zhang2024actuator}
S.~Zhang, J.~Zhu, T.-H. Huang, S.~Yu, J.~S. Huang, I.~Lopez-Sanchez, T.~Devine, M.~Abdelhady, M.~Zheng, T.~C. Bulea \emph{et~al.}, ``Actuator optimization and deep learning-based control of pediatric knee exoskeleton for community-based mobility assistance,'' \emph{Mechatronics}, vol.~97, p. 103109, 2024.

\bibitem{ahn2022can}
M.~Ahn, A.~Brohan, N.~Brown, Y.~Chebotar, O.~Cortes, B.~David, C.~Finn, C.~Fu, K.~Gopalakrishnan, K.~Hausman \emph{et~al.}, ``Do as i can, not as i say: Grounding language in robotic affordances,'' \emph{arXiv preprint arXiv:2204.01691}, 2022.

\bibitem{wu2023tidybot}
J.~Wu, R.~Antonova, A.~Kan, M.~Lepert, A.~Zeng, S.~Song, J.~Bohg, S.~Rusinkiewicz, and T.~Funkhouser, ``Tidybot: Personalized robot assistance with large language models,'' \emph{Autonomous Robots}, vol.~47, no.~8, pp. 1087--1102, 2023.

\bibitem{lynch2023interactive}
C.~Lynch, A.~Wahid, J.~Tompson, T.~Ding, J.~Betker, R.~Baruch, T.~Armstrong, and P.~Florence, ``Interactive language: Talking to robots in real time,'' \emph{IEEE Robotics and Automation Letters}, 2023.

\bibitem{ao2025llm}
J.~Ao, F.~Wu, Y.~Wu, A.~Swiki, and S.~Haddadin, ``Llm-as-bt-planner: Leveraging llms for behavior tree generation in robot task planning,'' in \emph{2025 IEEE International Conference on Robotics and Automation (ICRA)}.\hskip 1em plus 0.5em minus 0.4em\relax IEEE, 2025, pp. 1233--1239.

\bibitem{zhou2024isr}
Z.~Zhou, J.~Song, K.~Yao, Z.~Shu, and L.~Ma, ``Isr-llm: Iterative self-refined large language model for long-horizon sequential task planning,'' in \emph{2024 IEEE International Conference on Robotics and Automation (ICRA)}.\hskip 1em plus 0.5em minus 0.4em\relax IEEE, 2024, pp. 2081--2088.

\bibitem{liu2023task}
W.~Liu, X.~Liang, and M.~Zheng, ``Task-constrained motion planning considering uncertainty-informed human motion prediction for human--robot collaborative disassembly,'' \emph{IEEE/ASME Transactions on Mechatronics}, vol.~28, no.~4, pp. 2056--2063, 2023.

\bibitem{cui2023no}
Y.~Cui, S.~Karamcheti, R.~Palleti, N.~Shivakumar, P.~Liang, and D.~Sadigh, ``No, to the right: Online language corrections for robotic manipulation via shared autonomy,'' in \emph{Proceedings of the 2023 ACM/IEEE International Conference on Human-Robot Interaction}, 2023, pp. 93--101.

\bibitem{shi2024yell}
L.~X. Shi, Z.~Hu, T.~Z. Zhao, A.~Sharma, K.~Pertsch, J.~Luo, S.~Levine, and C.~Finn, ``Yell at your robot: Improving on-the-fly from language corrections,'' \emph{arXiv preprint arXiv:2403.12910}, 2024.

\bibitem{zha2024distilling}
L.~Zha, Y.~Cui, L.-H. Lin, M.~Kwon, M.~G. Arenas, A.~Zeng, F.~Xia, and D.~Sadigh, ``Distilling and retrieving generalizable knowledge for robot manipulation via language corrections,'' in \emph{2024 IEEE International Conference on Robotics and Automation (ICRA)}.\hskip 1em plus 0.5em minus 0.4em\relax IEEE, 2024, pp. 15\,172--15\,179.

\bibitem{jiang2024transic}
Y.~Jiang, C.~Wang, R.~Zhang, J.~Wu, and L.~Fei-Fei, ``Transic: Sim-to-real policy transfer by learning from online correction,'' \emph{arXiv preprint arXiv:2405.10315}, 2024.

\bibitem{de2008atlas}
A.~De~Santis, B.~Siciliano, A.~De~Luca, and A.~Bicchi, ``An atlas of physical human--robot interaction,'' \emph{Mechanism and Machine Theory}, vol.~43, no.~3, pp. 253--270, 2008.

\bibitem{haddadin2016physical}
S.~Haddadin and E.~Croft, ``Physical human--robot interaction,'' in \emph{Springer Handbook of Robotics}.\hskip 1em plus 0.5em minus 0.4em\relax Springer, 2016, pp. 1835--1874.

\bibitem{farajtabar2024path}
M.~Farajtabar and M.~Charbonneau, ``The path towards contact-based physical human--robot interaction,'' \emph{Robotics and Autonomous Systems}, vol. 182, p. 104829, 2024.

\bibitem{jarrasse2012framework}
N.~Jarrass{\'e}, T.~Charalambous, and E.~Burdet, ``A framework to describe, analyze and generate interactive motor behaviors,'' \emph{PloS One}, vol.~7, no.~11, p. e49945, 2012.

\bibitem{mortl2012role}
A.~M{\"o}rtl, M.~Lawitzky, A.~Kucukyilmaz, M.~Sezgin, C.~Basdogan, and S.~Hirche, ``The role of roles: Physical cooperation between humans and robots,'' \emph{The International Journal of Robotics Research}, vol.~31, no.~13, pp. 1656--1674, 2012.

\bibitem{losey2018review}
D.~P. Losey, C.~G. McDonald, E.~Battaglia, and M.~K. O'Malley, ``A review of intent detection, arbitration, and communication aspects of shared control for physical human--robot interaction,'' \emph{Applied Mechanics Reviews}, vol.~70, no.~1, p. 010804, 2018.

\bibitem{hoffman2024inferring}
G.~Hoffman, T.~Bhattacharjee, and S.~Nikolaidis, ``Inferring human intent and predicting human action in human--robot collaboration,'' \emph{Annual Review of Control, Robotics, and Autonomous Systems}, vol.~7, 2024.

\bibitem{losey2022physical}
D.~P. Losey, A.~Bajcsy, M.~K. O'Malley, and A.~D. Dragan, ``Physical interaction as communication: Learning robot objectives online from human corrections,'' \emph{The International Journal of Robotics Research}, vol.~41, no.~1, pp. 20--44, 2022.

\bibitem{li2021learning}
M.~Li, A.~Canberk, D.~P. Losey, and D.~Sadigh, ``Learning human objectives from sequences of physical corrections,'' in \emph{2021 IEEE International Conference on Robotics and Automation (ICRA)}.\hskip 1em plus 0.5em minus 0.4em\relax IEEE, 2021, pp. 2877--2883.

\bibitem{mehta2024unified}
S.~A. Mehta and D.~P. Losey, ``Unified learning from demonstrations, corrections, and preferences during physical human--robot interaction,'' \emph{Journal of Human-Robot Interaction}, vol.~13, no.~3, p.~39, 2024.

\bibitem{xie2024safe}
Z.~Xie, W.~Zhang, Y.~Ren, Z.~Wang, G.~J. Pappas, and W.~Jin, ``Safe mpc alignment with human directional feedback,'' \emph{arXiv preprint arXiv:2407.04216}, 2024.

\bibitem{losey2017trajectory}
D.~P. Losey and M.~K. O'Malley, ``Trajectory deformations from physical human--robot interaction,'' \emph{IEEE Transactions on Robotics}, vol.~34, no.~1, pp. 126--138, 2017.

\bibitem{cheng2019comprehensive}
G.~Cheng, E.~Dean-Leon, F.~Bergner, J.~R.~G. Olvera, Q.~Leboutet, and P.~Mittendorfer, ``A comprehensive realization of robot skin: Sensors, sensing, control, and applications,'' \emph{Proceedings of the IEEE}, vol. 107, no.~10, pp. 2034--2051, 2019.

\bibitem{mittendorfer2011humanoid}
P.~Mittendorfer and G.~Cheng, ``Humanoid multimodal tactile-sensing modules,'' \emph{IEEE Transactions on Robotics}, vol.~27, no.~3, pp. 401--410, 2011.

\bibitem{armleder2022interactive}
S.~Armleder, E.~Dean-Leon, F.~Bergner, and G.~Cheng, ``Interactive force control based on multimodal robot skin for physical human--robot collaboration,'' \emph{Advanced Intelligent Systems}, vol.~4, no.~2, p. 2100047, 2022.

\bibitem{magrini2014estimation}
E.~Magrini, F.~Flacco, and A.~De~Luca, ``Estimation of contact forces using a virtual force sensor,'' in \emph{2014 IEEE/RSJ International Conference on Intelligent Robots and Systems (IROS)}.\hskip 1em plus 0.5em minus 0.4em\relax IEEE, 2014, pp. 2126--2133.

\bibitem{manuelli2016localizing}
L.~Manuelli and R.~Tedrake, ``Localizing external contact using proprioceptive sensors: The contact particle filter,'' in \emph{2016 IEEE/RSJ International Conference on Intelligent Robots and Systems (IROS)}.\hskip 1em plus 0.5em minus 0.4em\relax IEEE, 2016, pp. 5062--5069.

\bibitem{zwiener2018contact}
A.~Zwiener, C.~Geckeler, and A.~Zell, ``Contact point localization for articulated manipulators with proprioceptive sensors and machine learning,'' in \emph{2018 IEEE International Conference on Robotics and Automation (ICRA)}.\hskip 1em plus 0.5em minus 0.4em\relax IEEE, 2018, pp. 323--329.

\bibitem{tian2023optimization}
S.~Tian, X.~Liang, and M.~Zheng, ``An optimization-based human behavior modeling and prediction for human-robot collaborative disassembly,'' in \emph{2023 American Control Conference (ACC)}.\hskip 1em plus 0.5em minus 0.4em\relax IEEE, 2023.

\bibitem{liu2024integrating}
W.~Liu, K.~Eltouny, S.~Tian, X.~Liang, and M.~Zheng, ``Integrating uncertainty-aware human motion prediction into graph-based manipulator motion planning,'' \emph{IEEE/ASME Transactions on Mechatronics}, vol.~29, no.~4, pp. 3128--3136, 2024.

\end{thebibliography}

\end{document}